\documentclass[letterpaper, 10 pt, conference]{ieeeconf}

\IEEEoverridecommandlockouts

\usepackage{hyperref}
\usepackage{graphics} 
\usepackage{epsfig} 
\usepackage{mathptmx} 
\usepackage{times} 
\usepackage{amsmath} 
\usepackage{amssymb}  
\usepackage{color}
\usepackage[ruled]{algorithm2e}
\usepackage{booktabs}
\usepackage{subcaption}
\usepackage[utf8]{inputenc}
 
\title{\LARGE \bf
Adaptive Genomic Evolution of Neural Network Topologies (AGENT) for State-to-Action Mapping in Autonomous Agents\thanks{Copyright\copyright 2019 IEEE. Personal use of this material is permitted.
  Permission from IEEE must be obtained for all other uses, in any current or future 
  media, including reprinting/republishing this material for advertising or promotional
  purposes, creating new collective works, for resale or redistribution to servers or 
  lists, or reuse of any copyrighted component of this work in other works. \vspace{0.2cm}}
}

\author{Amir Behjat$^{1}$ Sharat Chidambaran$^{2}$ Souma Chowdhury$^{3}$
\thanks{$^{1}$PhD. Student, Dept. of Mechanical \& Aerospace Engineering,
        University at Buffalo, Buffalo, NY 14260, USA.
        {\tt\small amirbehj@buffalo.edu}}%
\thanks{$^{2}$M.S Student, Dept. of Mechanical \& Aerospace Engineering,
        University at Buffalo, Buffalo, NY 14260, USA.
        {\tt\small sharatpa@buffalo.edu}}%
\thanks{$^{3}$Assistant Professor, Dept. of Mechanical \& Aerospace Engineering,
        University at Buffalo, Buffalo, NY 14260, USA. \textit{Corresponding Author}.
        {\tt\small soumacho@buffalo.edu}}%
}
\begin{document}
\maketitle
\begin{abstract}
Neuroevolution is a process of training neural networks (NN) through an evolutionary algorithm, usually to serve as a state-to-action mapping model in control or reinforcement learning-type problems. This paper builds on the Neuro Evolution of Augmented Topologies (NEAT) formalism that allows designing topology and weight evolving NNs. Fundamental advancements are made to the neuroevolution process to address premature stagnation and convergence issues, central among which is the incorporation of automated mechanisms to control the population diversity and average fitness improvement within the neuroevolution process. Insights into the performance and efficiency of the new algorithm is obtained by evaluating it on three benchmark problems from the Open AI platform and an Unmanned Aerial Vehicle (UAV) collision avoidance problem.
\end{abstract}
\section{Introduction}
\label{sec:intro}
Artificial Neural networks or ANNs are playing an emerging role as decision-support models in various intelligent autonomous systems \cite{dounis2009advanced}.
This emergence is partly attributed to the capability of ANNs to serve as universal function approximators, allowing them to be used for mapping states to (discrete or continuous) actions in autonomous systems. 
A significant fraction of such applications fall in the category where optimum actions corresponding to various states (i.e., labeled data) are not known apriori. Outside of classical control and planning methods, reinforcement learning (RL) methods \cite{c1.1,c1.2} and its recent deep variants \cite{c1.3} constitute a dominant player in training state-to-action models in such scenarios. However, RL methods use gradient information for back propagation which is not easy to ascertain for some problems \cite{grathwohl2017backpropagation},
and generally do not scale well with the dimension of the problem for many cases \cite{poggio2017and}. Most RL variants (with recent exceptions \cite{c1.4,liu2018parallel}) are also not conducive to application in the continuous-space domain. An alternative class of frameworks, based on evolutionary algorithms \cite{floreano2008neuroevolution}, namely neuroevolution \cite{c1.7} and evolution strategies \cite{hansen2003reducing}, seek to mitigate these limitations.

While neuroevolution allows highly parallelized implementations and can be applied to problems with continuous/mixed state spaces, they are often plagued with poor convergence, premature stagnation, and topological inflexibility issues. In this paper, we present a novel neuroevolution approach, with the aim to address these key issues.

Neuroevolution is the process of designing ANNs through evolutionary optimization algorithms. 
Early approaches merely evolved the weights of the ANN without altering its topology \cite{c1.8}, a typical continuous optimization problem -- genetic algorithms (GA) were used for this purpose. In these approaches, the architecture or topology of the ANN was user-prescribed (as is common across most domains of ANN training and usage), which however leads to sub-optimal of overfitting prediction models \cite{turner2013importance} for state-to-action mapping. Later endeavors introduced the concept of \textbf{\textit{topology and weight evolving ANNs}} (TWEANNs). Neuroevolution of augmenting topologies or NEAT \cite{c1.9} is perhaps the most well-known implementation of the TWEANN concept. NEAT evolves ANNs via a GA that directly encodes the nodes, edges, and edge weights (the phenotype) in a specialized genotype. NEAT commences with a population of minimalist genomes, represented by feedforward ANNs (with no hidden nodes), whose input and output layers are sized according to the problem at hand. At every generation of NEAT, along with standard genetic operations (i.e., selection, crossover, and mutation) a specialized operation called ``speciation" is performed on the population in order to preserve newly created complex topologies (with likely premature weights). Variations of NEAT, including HyperNEAT \cite{c1.10} (which provides an indirect genotype$\rightarrow$phenotype encoding) and SUNA \cite{c3.1} have been used to control virtual agents in Atari games \cite{c1.11}, evolve robot gaits \cite{c1.12}, geological prediction \cite{wang2013application}, and financial market analysis \cite{nadkarni2018combining}. More recently, neuroevolution has also been employed to evolve deep neural networks \cite{c1.13}. 

A persistent concern in evolving neural network topologies (for highly non-linear state$\rightarrow$action mapping) is that of premature stagnation. Neuroevolution methods are often unable to preserve genomic diversity as (nascent) complex structures cannot stabilize their weights as fast as simple networks leading to local stagnation (fitness functions typically tend to be highly non-convex over the NN topological space). While problem-specific tedious heuristics can strive to address this concern, \textit{we hypothesize that situation-adaptive \textbf{automated} variation of selection pressure and reproduction operators are needed to offer generalized solutions}. In addition, while the concept in NEAT of initializing the population with minimalist NN topologies \cite{c1.9} mitigates the possibility of overfitting (to sample scenarios used for fitness evaluation) and aids fast convergence for problems with low-dimensional input spaces, a converse effect is encountered in problems with larger state spaces or when modeling highly non-linear state$\rightarrow$action mapping. In these cases, starting with a minimalist baseline could lead to wasted computational effort to reach network topologies of reasonable complexity. \textit{Problem size-adaptive initialization of the NN topologies and allowing topological complexity to both increase and decrease \cite{james2004comparative} during neuroevolution is hypothesized to address this issue.} 

To investigate the above-stated hypotheses, this paper develops an \textit{adaptive neuroevolution} approach that incorporates novel mechanisms for insitu control of the genomic diversity and average fitness improvement. Further performance and computational efficiency gains are accomplished by allowing flexible topology initialization and provisioning nodes with variable activation function and memory properties. The new algorithm is evaluated both over benchmark RL and control problems and a practical robotics problem -- collision avoidance in unmanned aerial vehicles (UAV). The next section describes the basic components of the algorithm. Section \ref{sec: adaptive neuro} presents the salient features of the algorithm. Section \ref{sec:openai} and Section \ref{sec: uav-coll} respectively demonstrate the capabilities of the algorithm on benchmark problems and the UAV problem; and Section \ref{sec: concl} provides concluding remarks. 
\vspace{-0.1cm}

\section{AGENT (Neuroevolution) Algorithm}
\label{sec:main neuro}
\vspace{-0.1cm}
The new neuroevolution algorithm is called Adaptive Genomic Evolution of Neural Network Topologies or \textbf{AGENT}. In this section, we describe the key components of this algorithm, namely the intra-generational stages, the encoding approach, and the selection and reproduction operators.

\subsection{AGENT: Stages}
\begin{figure}[h]
\centering
\includegraphics[width=0.99\columnwidth]{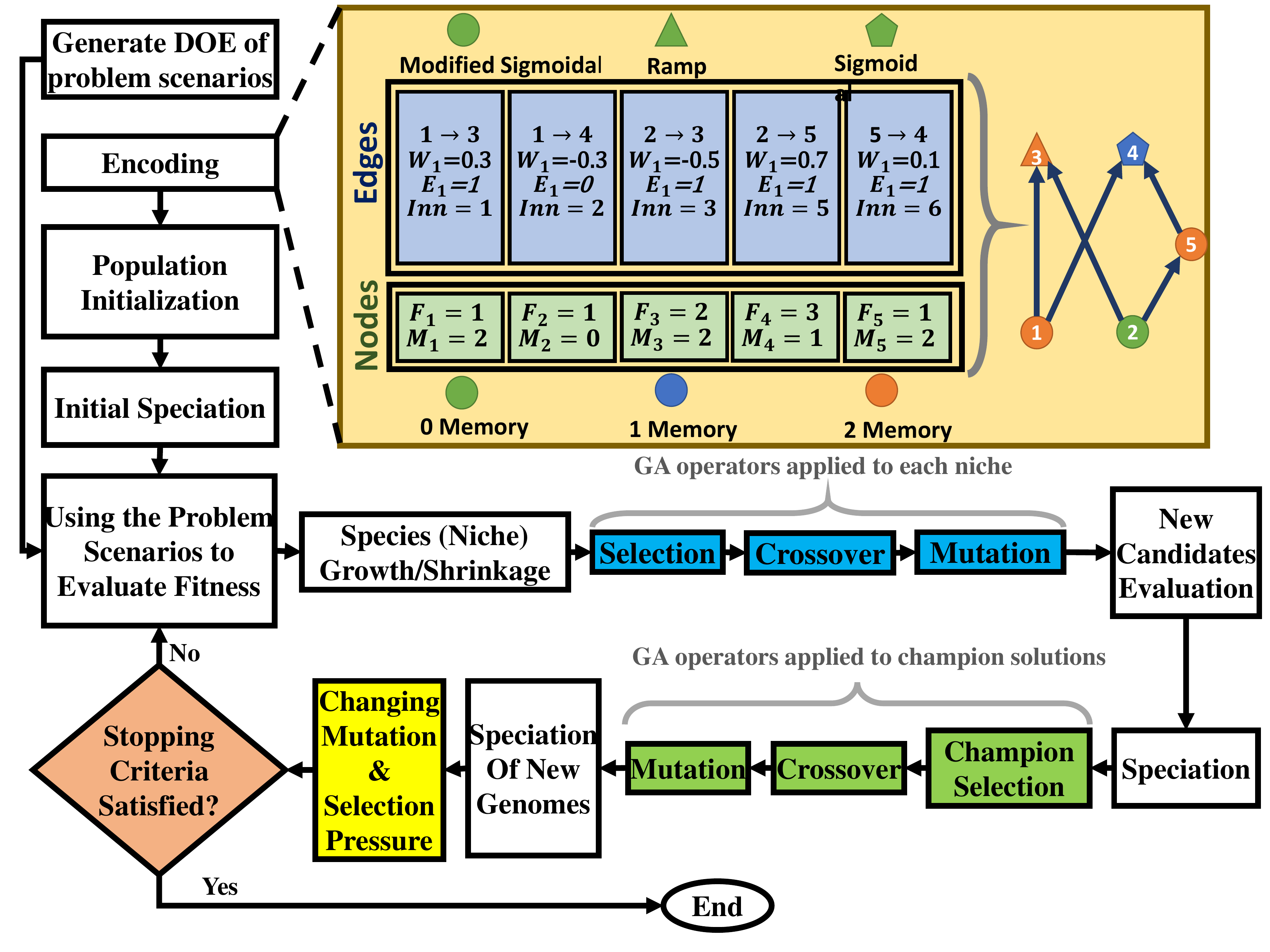}
\vspace{-0.3cm}
\caption{AGENT: Flowchart}
\vspace{-0.4cm}
\label{fig:flow_all}
\end{figure}

Figure \ref{fig:flow_all} illustrates the overall flowchart of the AGENT algorithm. AGENT uses a two-stage evolutionary approach. All species participate in the first stage of evolution in each generation, while only the best genomes from each species participate in the second stage of evolution. 

\subsection{Encoding}
Information-processing capacity in an ANN is encapsulated both in its nodes and edges. Therefore, similar to the original NEAT, AGENT uses a direct bi-structural encoding, where each genome comprises of a node encoding and an edge encoding. Where AGENT differs from NEAT is in how nodes are encoded -- in AGENT, the node encoding also defines the type of activation function and memory capacity of the node. To allow greater flexibility, one of three activation functions can be selected: modified sigmoid, (only option in original NEAT), saturated linear, and sigmoid functions. The \textit{memory} is allowed to take one of three values: $M\in\{0,1,2\}$; a memory size of $0$ designates using the current weighted input incoming into the node; memory sizes of $1$ and $2$ respectively allow using the first and second temporal derivatives of the weighted input incoming into the node. The latter enables exibiting temporal dynamic behavior, useful for neurocontroller type applications. 
Equation \ref{eq:memory} explains how the derivatives of each node are used. In this equation, for a node connected to $n_i$ upstream nodes, $V_{i}(\tau)$ is the net synaptic input of node-$i$ in time step $\tau$, $f_j$ is the output of the (upstream) node-$j$, $\delta \tau$ is the time step used when implementing this NN as a controller, and $U_{i}(\tau) = \sum_{j=1}^{n_i} (w_{j,i} \times f_{j})$.
\vspace{-0.1cm}
\begin{equation}
V_{i}(\tau) =
\begin{cases}
    U_{i}(\tau),  & {\rm{if}}~ M = 0\\
    \dfrac{U_{i}(\tau)-U_{i}(\tau-1)}{\delta \tau}, & {\rm{if}}~ M=1  \\ 
    \dfrac{U_{i}(\tau)-2 U_{i}(\tau-1)+ U_{i}(\tau-2)}{(\delta \tau)^2}, & {\rm{if}}~ M=2   \\
\end{cases}
\label{eq:memory}
\end{equation}
%

\subsection{Initialization}
Unlike NEAT, the initial population in AGENT is allowed to comprise a small number of hidden neurons, instead of a minimalist topology with no hidden neurons. 
To introduce diversity in the initial population, the number of hidden nodes in each genome is chosen from a distribution such that the expected value (over the population) of the number of hidden nodes is given by $\sqrt[]{n_I \times n_O}$; 
%
%
here $n_I$ and $n_O$ are respectively the number of input and output nodes.

\subsection{Speciation}
Speciation is a crucial aspect of AGENT -- it refers to the process in which the population is divided into several subgroups. Speciation is carried out for two reasons. First, it shelters newly generated genomes (containing yet-to-be-stabilized weights) from getting eliminated due to selection pressure. Second, it facilitates greater local search within each species.
The speciation process adopted here is similar to that in SUNA \cite{c3.1}. The most unique genomes are chosen to represent the different species. The remaining genomes are then classified into these species/groups based on their similarity to the aforementioned unique genomes.


\subsection{Selection}
Tournament selection is used in AGENT due to its property of being invariant in order-preserving transformations. Moreover, compared to proportional selection, tournament selection has the added benefit of being able to modify selection pressure by varying the ratio between the number of genomes that participate in the tournament and the number of genomes that are allowed to win the tournament. This will be used later on for controlling the selection pressure.

\subsection{Crossover}
\begin{figure}[htbp]
\centering
\includegraphics[width =0.6\columnwidth]{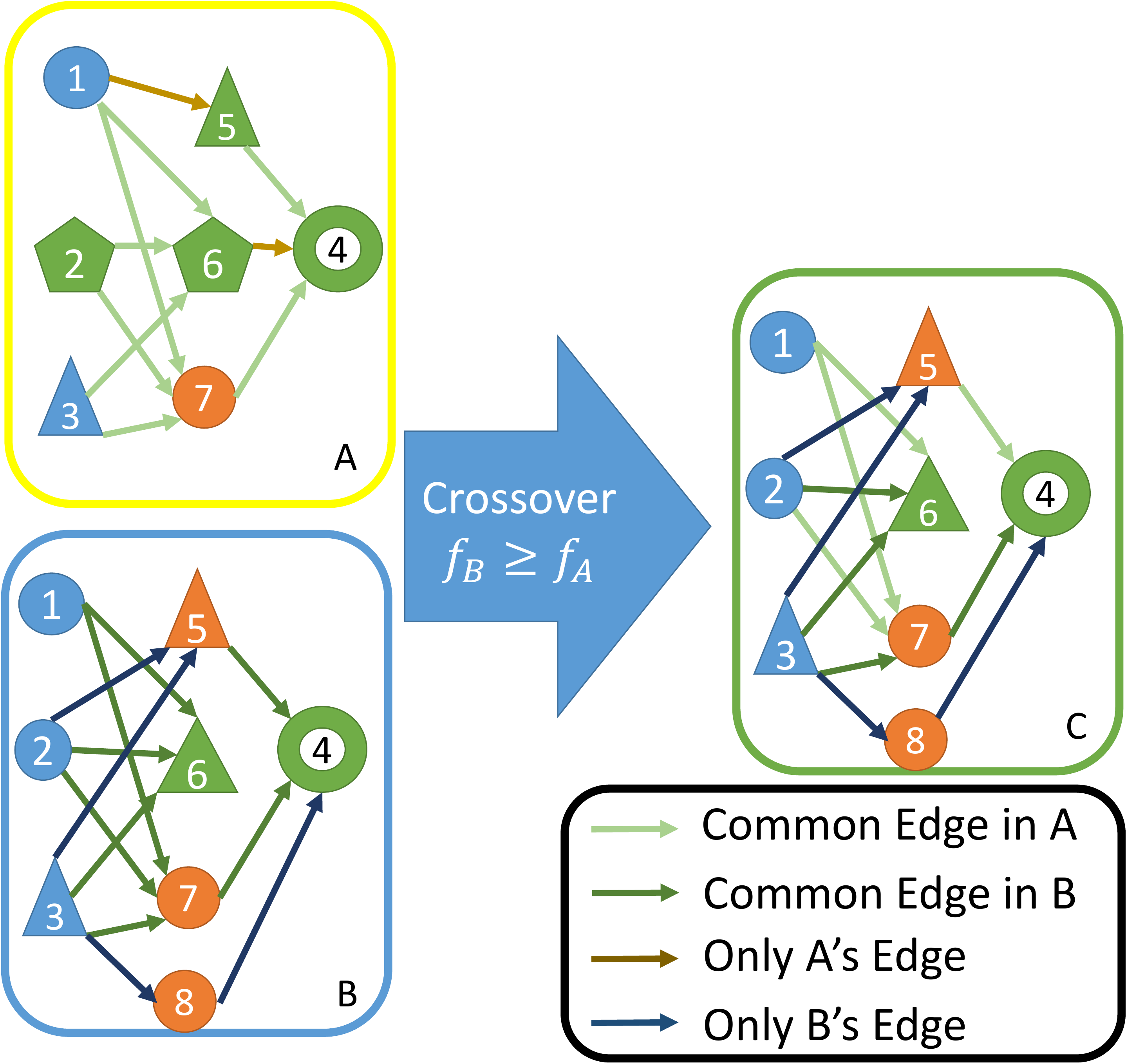}
 \captionsetup{justification=centering}
\caption{Crossover operation in AGENT}
\label{fig:Crossover}
\vspace{-0.6 cm}
\end{figure}

AGENT expands on the crossover process used in NEAT, by also transferring the special nodal properties (activation function and memory type). Figure \ref{fig:Crossover} illustrates this procedure. Weights of common edges are inherited randomly (with equal probability) from one of the two parents, while all nodal properties and weights associated with unique edges are inherited from the parent with superior fitness value.

\subsection{Mutation}
\begin{figure}[htbp]
\vspace{0.2cm}
\centering
\begin{tabular}{c}
\includegraphics[width =0.65\columnwidth]{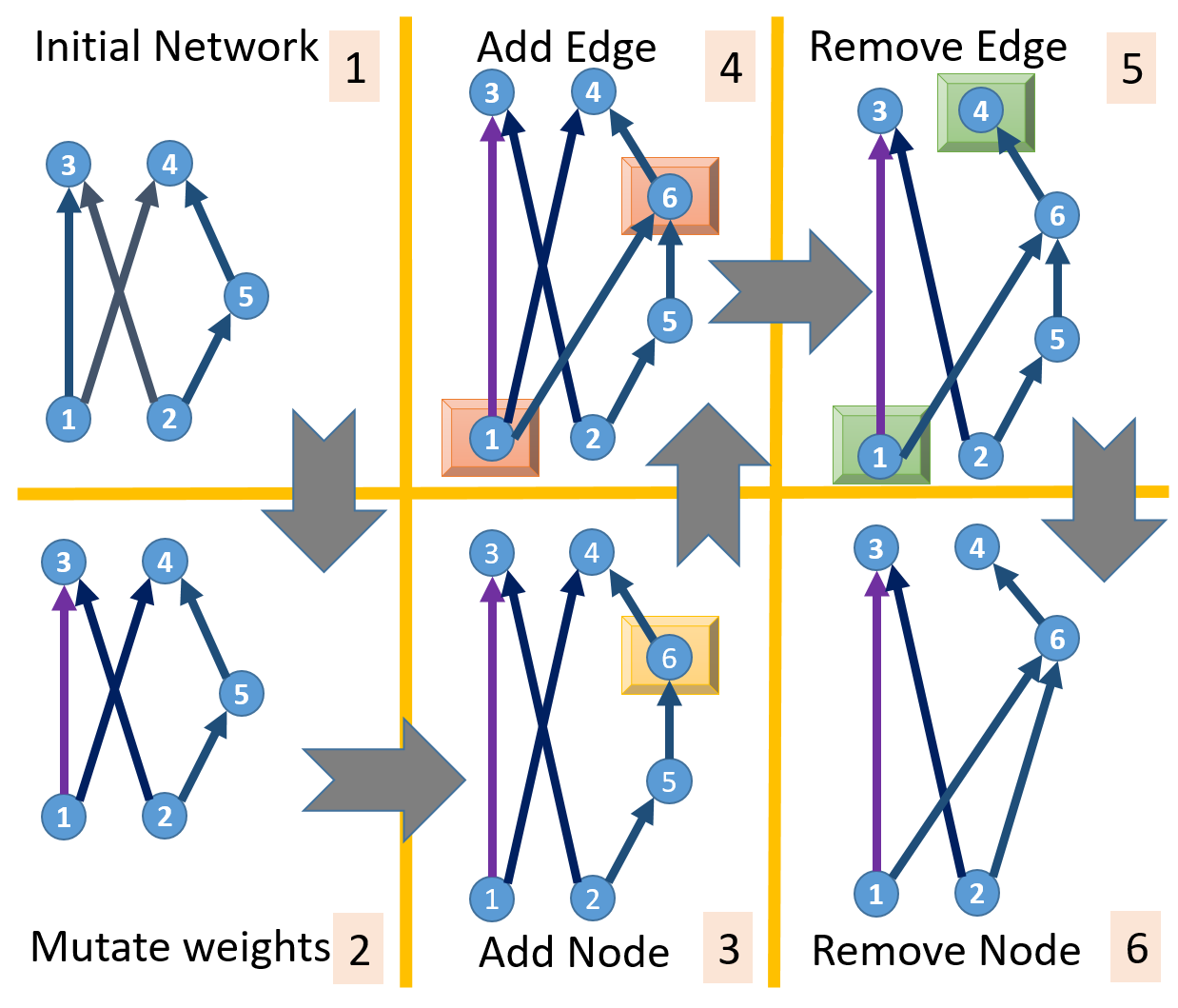}\\
(a) Mutation of edges and nodes\\
\includegraphics[width =0.8\columnwidth]{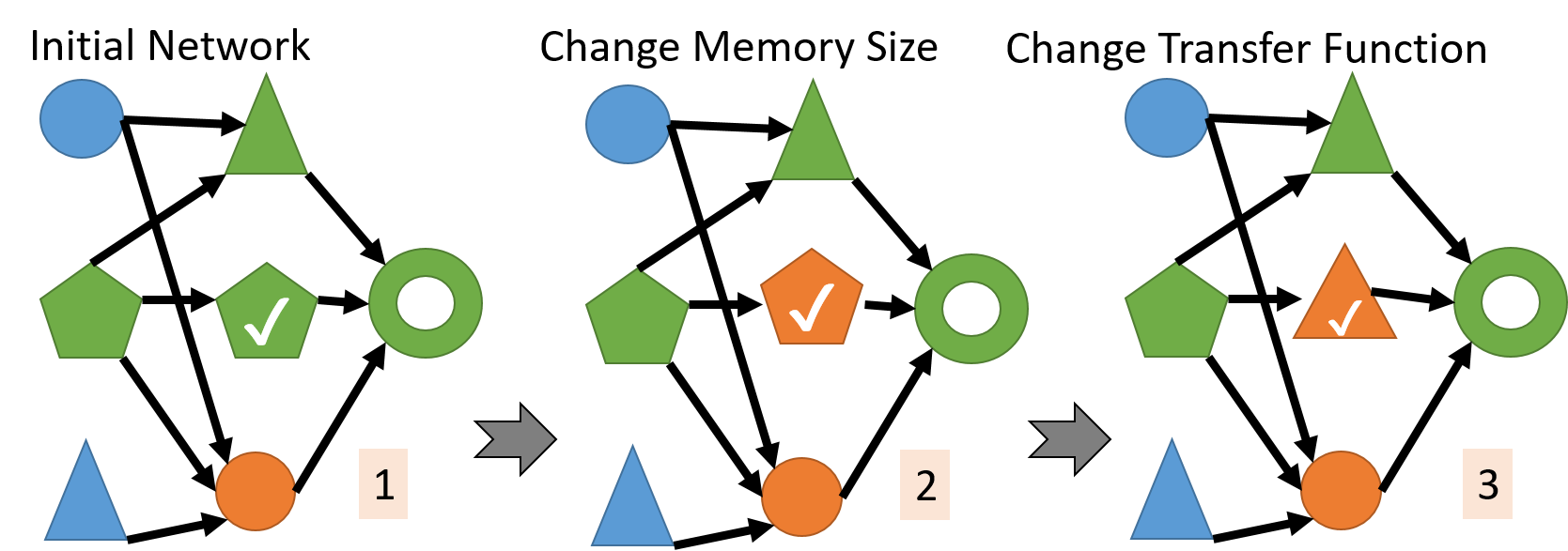}\\
(b) Mutation of nodal properties
\end{tabular}
 \captionsetup{justification=centering}
\caption{Mutation operations in AGENT}
\label{fig:Mutations}
\vspace{-0.5 cm}
\end{figure}

As illustrated in Fig. \ref{fig:Mutations} and described below, there are two types of mutation in AGENT, mutation of edges/nodes and mutation of nodal properties, each with their own rate.

\textit{Mutation of edge weights}:
For existing edges between any two nodes $i$ and $j$, real-valued Gaussian mutation \cite{Deb_book} of weights is undertaken, as given by:
\begin{equation}
w_{i,j} = w_{i,j} +r \in \mathcal{N}(0,\sigma_W)
\label{eq:mutation weight}
\end{equation}

\textit{Addition of an edge}: 
An edge can be added between any existing pair of nodes as long as duplicate edges and cycles are not produced. It must be noted that the addition of an edge/node increases the complexity of the network. The weight of the new edge is assigned randomly from a uniform  distribution in the range $[-1,+1]$. 

\textit{Removal of an edge}:
Removing edges and nodes assists in reducing network complexity, e.g., to mitigate overfitting. 
An edge between any existing pair of nodes can be removed as long it does not result in a floating node. 
In this research, the mutation rate for removing an edge is kept at 80$ \%$ of the mutation rate for adding an edge, thereby allowing slightly greater probability of network complexification. 

\textit{Addition of a node}:
A node can be added between any edge resulting in the splitting of the edge into two edges. 

\textit{Removal of a node}:
Any hidden node can be removed. Upon removal of a node, new connections are made such that all incident downstream nodes (w.r.t. to the removed node) are connected to all upstream nodes (to which the removed node was connected). 
The probability of removing a node is kept at $80\%$ of the probability of adding a node. 

\textit{Mutation of nodal properties}:
This is done by probabilistic switching between the categorical values of these properties (i.e., different memory value or activation function). 


\section{Adaptation Mechanisms in AGENT}\label{sec: adaptive neuro}
In this section, we outline the novel formulations proposed to adaptively control the diversity and (fitness) improvement rate of AGENT along with the requisite metrics and limits.

\subsection {Diversity Preservation: Measure of Diversity}\label{subsec:dis_metric}

Population diversity is paramount to effective neuroevolution. This calls for prudently controlling the population diversity -- an abrupt decrease in diversity can lead to premature stagnation, but at the same time, a steady (low) rate of diversity reduction is needed for exploitation and eventual convergence. The first step in diversity preservation is robust measurement of diversity. To develop a diversity measure, an approach is needed to quantify the differences between any two neural networks in the design space. In neuroevolution, since genomes encode different topologies, their basic dimensionality varies across the population. Hence, a distance metric similar to the novelty metric described in \cite{c3.1} is used here. The distance between two candidate ANNs, $A$ and $B$, is thus given by the weighted sum of the difference between their node types, as well as the difference between edges connecting different types of nodes. 
\begin{equation}\label{eq:dist-metric}
d_{A,B}= \alpha_{P} \sum_{i=1}^{P_{T}} |P_{i,A}-P_{i,B}| + \alpha_{E} \sum_{i=1}^{P_{T}} \sum_{j=1}^{P_{T}} |E_{i \rightarrow j,A}-E_{i \rightarrow j,B}|
\end{equation}
In this equation, $P_{i,A}$ is the number of nodes of type $i$ in neural network A. $E_{i\rightarrow j,A}$ is the number of edges from node type $i$ to node type $j$ in neural network A, and $N_{T}$ is the number of types of activation functions allowed in the NNs. The weights $\alpha_{P}$ and $\alpha_{E}$ are prescribed to be $0.5$ in this paper. 



Now, to quantify the overall diversity in the population at any given generation $t$, we construct a complete undirected graph $K_N$ out of the population of $N$ candidate NNs. The length of the edges connecting candidate NNs in this graph is given by the above defined distance metric (Eq. \ref{eq:dist-metric}). Then, employing the concept of \textit{minimum spanning tree} (MST), the total length of the MST is used as the diversity metric ($D_t$) at the $t^{\rm{th}}$ generation, as given by: 
\begin{equation}
D_t= \sum_{\forall ~ A,B\in \rm{M.S.T}} d_{A,B}
\end{equation}
where $A,B$ represents an edge connecting ANNs A and B in the (population) graph. Kruskal`s Algorithm \cite{c3.2} is used to determine the MST, which is computationally inexpensive ($O\left(|K_N| \log |K_N|\right)$, where $|K_N|$ is the number of edges in the graph), and thus can be called in every generation.




    
        
        




With this measure of diversity, we can define a desired value for diversity and also delineate an approach to maintain this desired value. 
The below proposed limit defines the desired diversity at any $t^{\rm{th}}$ generation.
\begin{equation}
D_{d,t} = D_{Init} \times \frac{\beta_{Div} \times  t_{max}-t}{\beta_{Div} \times t_{max}}
\end{equation}
Here $D_{Init}$ is the diversity in the initial population.
The coefficient $\beta_{Div} \geq 1$ is used to increase the diversity. Here, $t_{max}$ is the maximum allowed generations. This formulation suggests a low steady linear decrease of diversity. 



\subsection {Diversity Preservation: Controlling Diversity}\label{ssec:diversity}
The tournament size in the selection operator is used as the control input to regulate the diversity. The probability of selecting a specific genome, to be copied into the mating pool, decreases with the number of genomes participating in the tournament and increases with the number of the genomes chosen from the tournament.

The probability of the $k^{\rm{th}}$ ranked genome to be selected into the mating pool ($\mathbb{P}$) by resampling is given by:
\begin{equation}
p(k \in \mathbb{P}) \approx \left(1-\binom{N_{\rm{T}}}{N_{\rm{W}}}\left(\frac{N-k}{N}  \right)^{N_{\rm{W}}}\right)^{\frac{2 \times N}{N_{\rm{W}}}}
\end{equation}
%
where $N$ is the population size, and $N_{\rm{T}}$ and $N_{\rm{W}}$ are respectively the numbers of genomes that enter the tournament and win the tournament. Since crossover in AGENT produces a single child NN from two parent NNs, the numerator is multiplied by $2$. Based on this formulation, it can be seen that the probability of choosing lower ranked genomes increases by increasing the ratio $\frac{N_{\rm{W}}}{N_{\rm{T}}}$. Therefore this ratio can be used to decrease the selection pressure, thereby increasing the diversity, and thus serves as a suitable choice for a control input. For regulating diversity at any $t^{\rm{th}}$ generation, this control input can be computed as:
\begin{equation}
\left.\frac{N_{\rm{W}}}{N_{\rm{T}}}\right|_t =\left.\frac{N_{\rm{W}}}{N_{\rm{T}}}\right|_{t-1} \times  e^{-K_{D} \left(D_{t}-D_{d,t}\right)}
\end{equation}
Here $\left(D_{t}-D_{d,t}\right)$ represents the difference between the observed diversity and the desired diversity; $K_{D}$ is the diversity gain coefficient, which modifies the amount of change that must be applied to the tournament ratio. For the current study, we used $K_{D} =0.1$.

\subsection{Improvement Adaptation: Metric of Improvement}
The premise behind tracking and controlling fitness is its ability to reflect whether adequate search dynamics is present in the population. Since, diversity is simultaneously being preserved (Section \ref{ssec:diversity}), steady improvement in average fitness over generations (in comparison to the improvement in the best fitness in the population) would be reflective of a robust search process. With this premise, we first define an improvement metric that encapsulates the history of improvement, as given by.
\begin{equation}\label{eq:improve}
I_{t}= \int_{0}^{t-1} \left(\alpha_I\left(f_{t}-f_{\tau}\right)\right)^{\frac{\tau}{t} }d\tau
\end{equation}
where $t$ is the current generation, $f_t$ and $f_{\tau}$ respectively represent fitness function values at the $t^{\rm{th}}$ and $\tau^{\rm{th}}$ generations, and $\alpha_I$ is an scaling coefficient. This metric is such designed that more recent improvements have a greater influence. 





\subsection{Improvement Adaptation: Mutation Controller}
If the improvement (over generations) in the average fitness of the population lags far behind the improvement in the best fitness value, it demonstrates a weakening search dynamics across the population. 
Now, in TWEANNs, mutation is the main driver of network innovation. So, too high a rate of mutation continues generating new niches of NNs that do not get time to stabilize their weights, and the algorithm starts acting as random search leading to the lagging average fitness improvement scenario mentioned above. This is where an adaptive reduction in mutation rate is needed. Conversely, when the improvement in the fitness of the population best starts lagging behind improvement in average fitness of the population, it is indicative of potential stagnation at local optima, and calls for increasing the mutation rate to facilitate discovery of new networks. With this perspective, we propose the following mutation rate ($\mu_t$) control strategy:
%
%
\begin{equation}\label{eq:mutation-controller}
\mu_{t} =\mu_{t-1} \times e^{-K_{I} \times \frac{I_{Best,t}-I_{Ave,t}}{I_{Best,t}}}
\end{equation}
%
Here $\mu_{t}$ is the mutation rate in generation $t$, and $K_{I}$ is the mutation controller gain coefficient. Similiar to $K_D$, the mutation controller gain can be prescribed to enable more or less aggressive search dynamics. For the current paper, $K_I$ is set at 0.1. In Eq. \ref{eq:mutation-controller}, $I_{Ave,t}$ and $I_{Best,t}$ respectively represent the fitness improvement metrics for the population average and the population best; they are computed using Eq. \ref{eq:improve}.


\section{Benchmark Testing of AGENT: OpenAI Gym}
\label{sec:openai}
OpenAI Gym is an open-source platform \cite{c4.1} that has been growing in popularity for benchmarking and comparing RL algorithms \cite{c1.5}, as well as other learning and optimization methods \cite{c1.13} that can solve RL-type control problems. 
In this paper, we showcase the performance of AGENT on three
problems curated from the OpenAI gym and compare with published results on state-of-the-art RL methods (summarized in Table \ref{table:compare_res}). These problems are very briefly described below; further information on these implementations, e.g., details of the state and action vectors, can be found at {\footnotesize\url{https://github.com/openai/gym/wiki/Leaderboard}}.

\subsection{Mountain Car}
In the Mountain Car problem, a candidate NN agent must control an underpowered car so that it can successfully climb up a mountain. 
In this paper, the \textit{MountainCarContinuous-v0} environment taken directly from OpenAI Gym is used; for the sake of fair comparison, the same reward function as described in the source code is used. 
The initial position of the car is randomly generated at the commencement of each episode -- this could lead to misleading results, as some solutions might accumulate a good reward due to a conducive starting position. To account for this factor, the number of episodes each genome encounters 
is controlled in a progressive manner. Each genome must accumulate reward thresholds before progressing on to the next episode. Mathematically, we express this as:
\begin{equation}
F_{i} = \sum_{j=1}^{N_{act,i}}{R_{i,j}}; ~~~ F_{net} = \frac{1}{N_{s}}\sum_{i = 1}^{N_p}{F_{i}}
\label{eq:accumulated reward}
\end{equation} 
%
%
where $R_{i,j}$ is the reward/penalty the agent receives for each action taken; $N_{act,i}$ is the total number of actions taken in the $i$-th scenario, and $F_{i}$ represents the genome's accumulated reward in that scenario; $F_{net}$ represents the net fitness function evaluate for a candidate genome; $N_{s}$ refers to the maximum number of scenarios available at training; and $N_p$ refers to the number of scenarios that the genome successfully passed. 

For this test problem, performance of AGENT is compared to that reported for the deep deterministic policy gradient method \cite{c4.4}. From the results in Table \ref{table:compare_res}, it can be seen that, AGENT was able to find significantly better reward values, albeit at the expense of a greater total number of steps (where a step is defined as an executed state$\rightarrow$action instance). 






\begin{table}[t]
\vspace{+.2cm}
\caption{OpenAI results: AGENT vs. Reference papers}
\label{table:compare_res}
\begin{tabular}{|l||c||c||c|}
\hline
\textbf{Parameter} & \textbf{Mount. Car}& \textbf{Acrobat} &  \textbf{Lunar} \\
\hline
 Population Size & 200  & 600 & 400\\
\hline
\textbf{AGENT: Best Reward} & \textbf{99.1} & \textbf{-69.6}& \textbf{68.0} \\
\hline
AGENT: Tot. Func. Eval. & 13,500  & 49,800 & 57,024\\
\hline
AGENT: Total Episodes  & 8,059 & 71,826 & 285,120\\
\hline
AGENT: Total Steps  & 855,199  & 73,831,050 & 1,369,446\\
\hline
\specialrule{.05em}{1em}{0em} 
 \textbf{Published Best Reward}  &  $\sim \textbf{90}  $ \cite{c4.4}& $ \sim \textbf{-70} $ \cite{c4.6} & $ \sim \textbf{200} $ \cite{c4.5}  \\
\hline
Published Total Episodes  & -  & 1,000& 10,000  \\
\hline
Published Total Steps  & 100,000 & - & - \\

\hline
\end{tabular}
\footnotesize{Published best values (under maximization) are taken from \cite{c4.4,c4.6,c4.5}.}
\vspace{-0.5cm}
\end{table}

\subsection{Acrobot}
The \textit{Acrobot-v1} environment in OpenAI Gym describes a two joint and two link robotic arm that initially hangs downwards. The goal in this problem is to produce a joint torque so as to swing the lower link up to a specified height.
The same approach as outlined in Eq. \ref{eq:accumulated reward} 
is taken to mitigate the effects of randomness in the environment. As can be seen from Table \ref{table:compare_res}, AGENT is able to achieve better reward values compared to that of the reference method, RL with adaptive memory replay \cite{c4.6}, again at the expense of additional computational cost.

\subsection{Lunar Lander}
The \textit{LunarLander-v2} environment in OpenAI Gym presents a problem with a mixture of discrete and continuous variables. Here, the agent must safely land a spacecraft on a launching pad via control of its three engines. 
The reward function described in OpenAI Gym is used. The effects of uncertainties in this problem (attributed to noisy engine thrust) are mitigated using the approach outlined in Eq. \ref{eq:accumulated reward}.

The optimum results obtained by AGENT is compared to that of a recently reported experience replay based RL method \cite{c4.5}. As seen from Table \ref{table:compare_res}, AGENT did not perform as well as the RL method. This can be attributed to the reward function for this problem, which presents many local minima, demanding different behaviors. Note that the handcrafted design of the reward functions in such problems do not necessarily represent generic performance in physical terms, and are often more amenable to RL based learning.

\begin{figure}[htbp]
\centering
\includegraphics[width=0.35\textwidth]{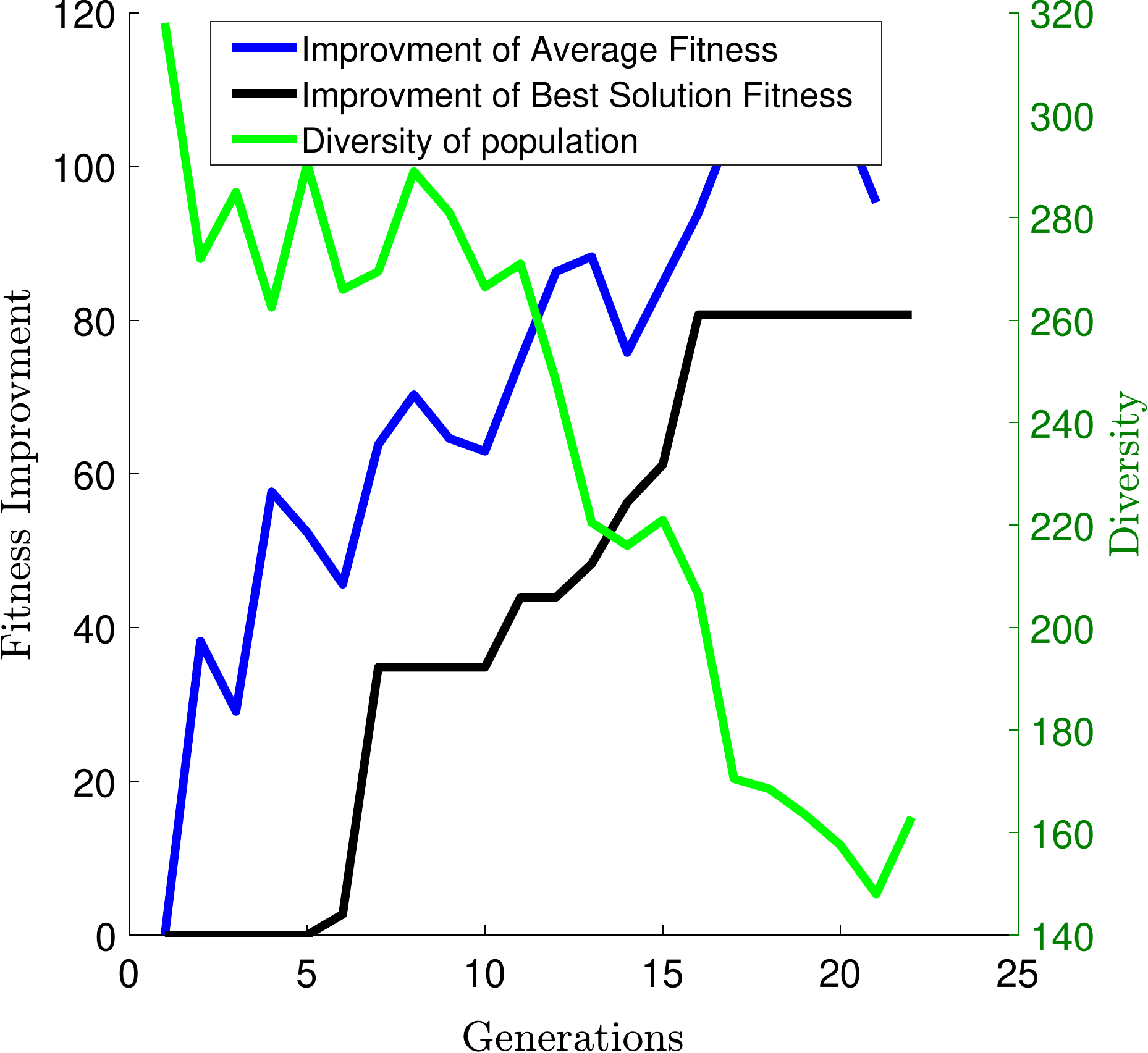}
\caption{Lunar Lander: diversity and mutation control effects}
\vspace{-.5cm}
\label{Figs:lunar_conv}
\end{figure}

Given this problem's complexity, it is used to analyze the performance of the special controllers in AGENT. 
Figure \ref{Figs:lunar_conv} shows the fitness improvement (not fitness value) of the population best and population average over generations, which are observed to follow similar trajectories $-$ thus providing evidence towards the effectiveness of the fitness improvement adaptation. A baseline case with the diversity/mutation controllers deactivated was also run, which provided an inferior optimum reward function of 47.4, further supporting the usefulness of AGENT's adaptation mechanisms.  
%

\section{Optimal UAV Collision Avoidance}
\label{sec: uav-coll}
In this section, we present the performance of AGENT on a UAV collision avoidance application. 
UAV collision avoidance is a well studied problem with solutions existing for avoiding both static \cite{iacono2018path} and dynamic \cite{kumar} obstacles. 

\begin{figure}[htbp]
\centering
\vspace{-0.3cm}
\includegraphics[page=1,width=0.99\columnwidth]{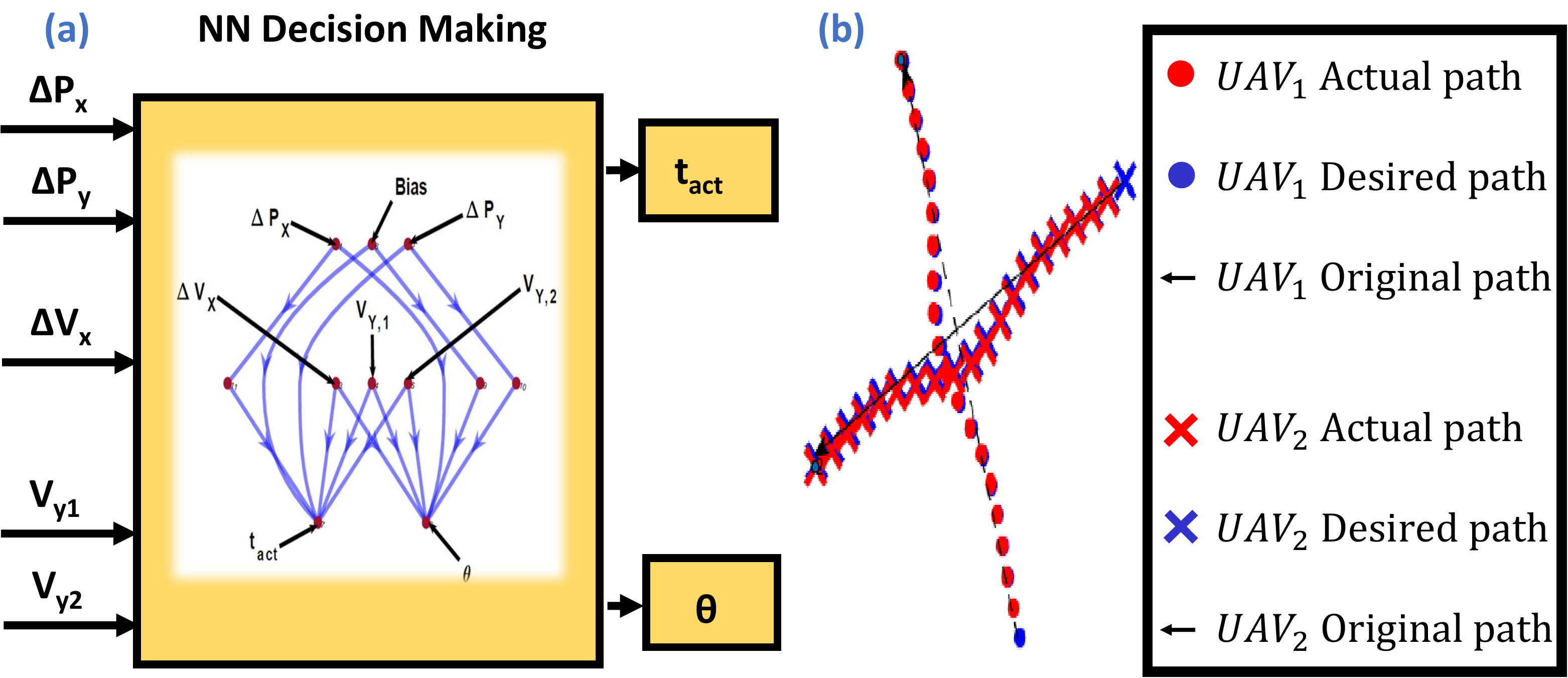}
\caption{UAV Collision Avoidance: (a) state$\rightarrow$action mapping, (b) Heading change maneuver is a given collision scenario
}
\vspace{-0.3cm}
\label{Fig:UAV_tri}
\end{figure}

The thesis \cite{c5.1} described an online cooperative collision avoidance approach for uniform quadcopter UAVs, where the UAVs undertake either a heading change or speed change maneuver, both in a reciprocal manner, e.g., if one UAV decides to veer to its left, the other UAV will also veer to its own left (both must get back to their original path). 
The prior approach used supervised learning, with optimization derived labels over sample collision scenarios, to train the maneuver models. Here, we use AGENT/neuroevolution to train the heading-change maneuver model. The outputs of the maneuver model are the time, $t_d$, between the point of detection and maneuver initiation, and the effective change in angle $\theta$. These are used to generate waypoints, then translated into a minimum snap trajectory, to be executed by a PID controller \cite{c5.1}. The inputs to the model include five UAV pose variables that completely define a collision scenario. 
Figure \ref{Fig:UAV_tri}(a) illustrates the inputs (state vector) and outputs (action vector) for this problem, and Fig. \ref{Fig:UAV_tri}(b) illustrates the online maneuver (given by the AGENT-trained model) for a representative collision scenario.  

In AGENT, each candidate genome is subjected to a set of $N_{s}$ collision scenarios, with the fitness function (to be maximized) given by the following aggregate performance:
\vspace{-0.2cm}
%
\begin{equation}
F_{CA} =1 - \frac{ E_{B} \sum_{i=1}^{N_{s}}   (\max(d_{d}-\min(g(d),d_{d}),0))^{2}  + E}{2 E_B}
\label{eq:UAV objective function}
\end{equation}
%
%
%
Here $E$ is the net energy consumed by both UAVs to execute the maneuver, and $E_{B}$ is the total battery capacity. In general, $E_B>>E$, and is used as a scaling factor to give more prominence to safety (i.e., maintaining adequate inter-UAV separation), compared to energy efficiency. In exceptional cases, where $E > E_B$ (typically indicates a diverging maneuver), the term $g(d)=0$, otherwise $g(d)$ is equal to the minimum separation distance experienced during the maneuver. 
The parameter $d_{d}$ represents the separation threshold, coming closer than which is considered as collision; $d_d$ is set at $1.5 m$.  

\begin{figure}[htbp]
\centering
\includegraphics[width=0.99\columnwidth]{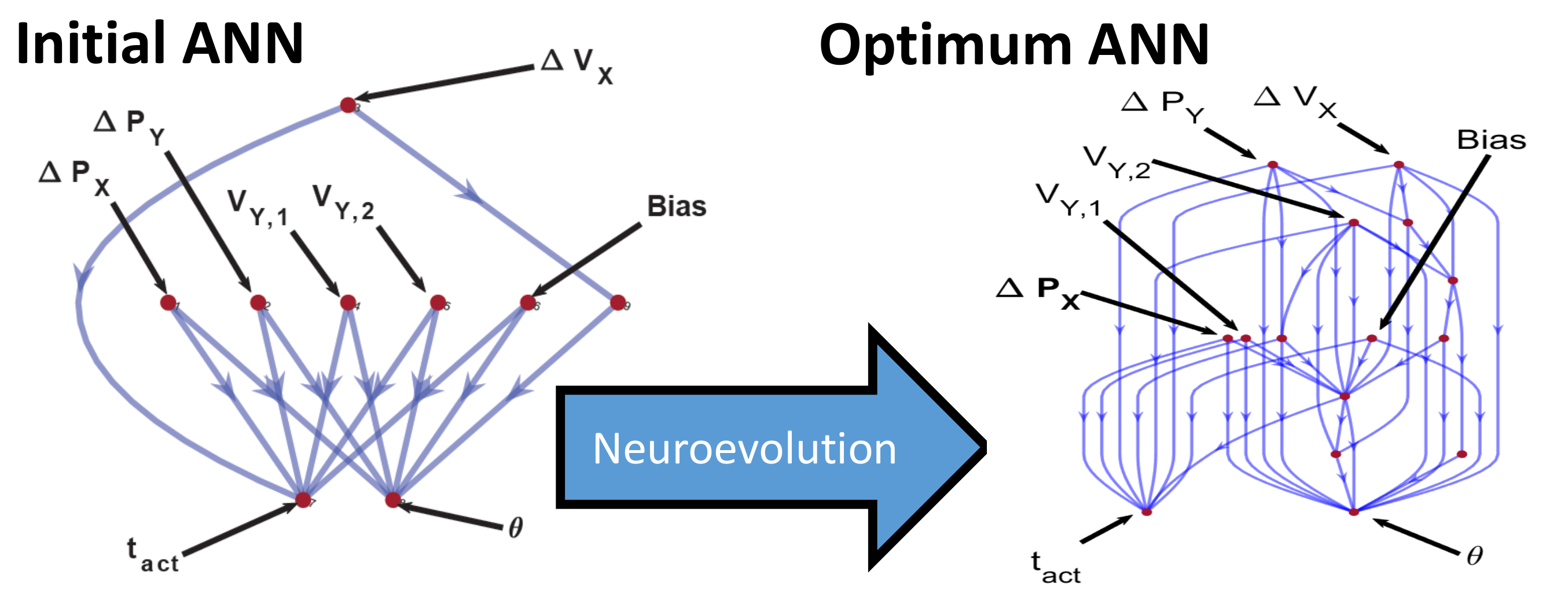}
\vspace{-0.4cm}
\caption{UAV problem: Evolution of NN via AGENT}
\label{Fig:UAV_NN}
\end{figure}

AGENT is run with a population size of 400 and allowed 60 generations. Figure \ref{Fig:UAV_NN} shows the structure of the optimum NN obtained, compared to an initial NN. This optimum NN avoids collisions in all $N_s=50$ training scenarios. 
It is also tested on an additional 200 unseen scenarios, chosen from the same distribution as the training scenarios. It successfully avoids collisions in 192/200 unseen scenarios. 

\begin{figure}[htbp]
\centering
\includegraphics[width=0.35\textwidth]{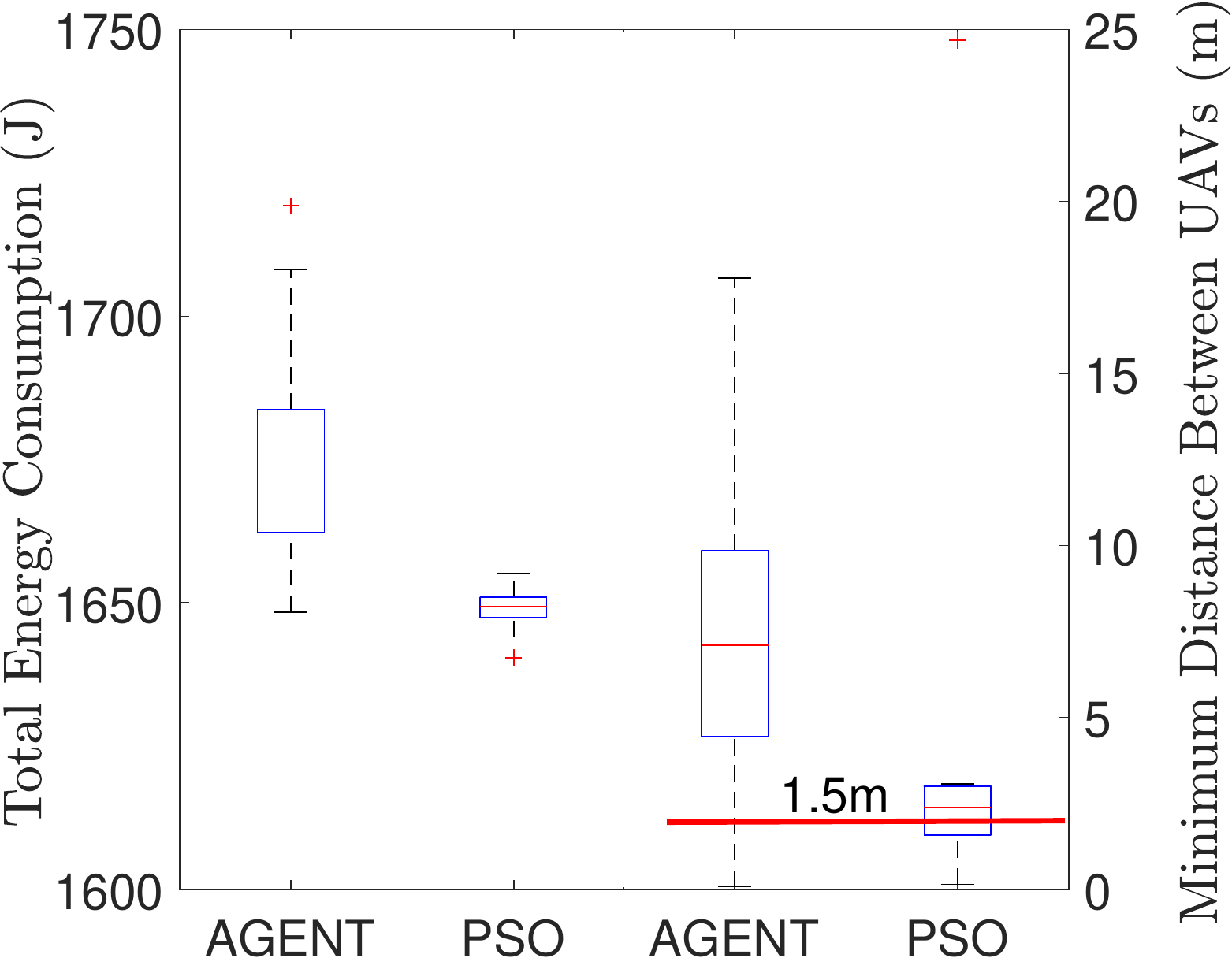}
\caption{UAV problem: Performance of AGENT compared to direct maneuver optimization over unseen test scenarios}
\vspace{-0.5cm}
\label{Fig:UAV_boxplot}
\end{figure}

For further validation, the performance of the AGENT-derived NN model is compared with that given by optimizing the maneuver individually for each test scenario. The PSO-based global optimization approach \cite{c5.2}, used in the prior work for generating labels \cite{c5.1}, is employed for the latter (whose computing cost makes it unsuitable for online application). Figure \ref{Fig:UAV_boxplot} shows the distribution of energy consumption and minimum separation distance (during maneuver) over the test scenarios, for both the AGENT-derived model and the PSO results. 
It can be observed from Fig. \ref{Fig:UAV_boxplot} that while the energy efficiency performance of AGENT's NN model is slightly worse than the PSO results, the avoidance performance is quite comparable -- 192/200 successes by AGENT's NN model vs. 196/200 successes by PSO.  

\section{Conclusion}
\label{sec: concl}
\vspace{-0.2cm}
In this paper, we developed a new neuroevolution method, called AGENT, by making important advancements to the NEAT formalism, The goal was to mitigate premature stagnation issues and improve the rate of convergence on complex RL/control problems. The key contributions included: 1) incorporating memory and activation function choice as nodal properties, 2) quantifying diversity using minimum spanning tree and controlling diversity via adaptive tournament selection, 3) controlling average fitness improvement via mutation rate adaptation, and 4) allowing both growth and shrinkage of NN topologies during evolution.  

The AGENT algorithm was tested on benchmark control problems adopted from the Open AI Gym, illustrating competitive results in terms of final outcomes (except in one problem), while incurring greater time steps cost, both compared to state-of-the-art RL methods. However, it is important to point out that AGENT is significantly more amenable to parallel implementation than RL methods, and thus computational time comparisons in future might elicit a different (likely more promising) picture. AGENT was also tested on an original UAV collision avoidance problem, resulting in an online model that provided competitive performance w.r.t offline optimization over test scenarios. Immediate future efforts will explore mechanisms to accelerate the evolutionary process via indirect genomic encoding and distributed implementations, in order to allow application to higher-dimensional learning problems in robotics.


\end{document}